\documentclass[conference]{IEEEtran}
\IEEEoverridecommandlockouts
\usepackage{CJKutf8}
\usepackage{tabularx}
\usepackage{cite}
\usepackage{amsmath,amssymb,amsfonts}
\usepackage{algorithmic}
\usepackage{graphicx}
\usepackage{textcomp}
\usepackage{xcolor}
\usepackage{bbm}
\usepackage{multirow}
\usepackage{array}
\usepackage{url}
\usepackage{booktabs}
\usepackage{array}
\usepackage{fancyhdr}

\def\BibTeX{{\rm B\kern-.05em{\sc i\kern-.025em b}\kern-.08em
    T\kern-.1667em\lower.7ex\hbox{E}\kern-.125emX}}

\begin{document}
\begin{CJK}{UTF8}{gbsn}
\title{RoKEPG: RoBERTa and Knowledge Enhancement for Prescription Generation of Traditional Chinese Medicine
}
\author{\IEEEauthorblockN{Hua Pu\textsuperscript{1}, 
		Jiacong Mi\textsuperscript{1}, 
		Shan Lu\textsuperscript{2}, 
		Jieyue He\textsuperscript{1,*}}\thanks{*Corresponding author}
		\IEEEauthorblockA{
		\textsuperscript{1}\textit{School of Computer Science and Engineering, Key Lab of
			Computer Network and Information Integration, MOE,}  \\
			\textit{Southeast University, Nanjing, 210018, Jiangsu, China}\\
		\textsuperscript{2}\textit{Nanjing Fenghuo Tiandi Communication Technology Co., Ltd, Nanjing, 211161, Jiangsu, China}} 
		\IEEEauthorblockA{Puhua2023@outlook.com, mijiacong@seu.edu.cn, bfcat.cn@gmail.com, jieyuehe@seu.edu.cn}
		}
\maketitle

\thispagestyle{fancy}

\lfoot{979-8-3503-3748-8/23/\$31.00 \copyright2023 IEEE}

\cfoot{}
\renewcommand{\headrulewidth}{0mm}
\begin{abstract}
Traditional Chinese medicine (TCM) prescription is the most critical form of TCM treatment, and uncovering the complex nonlinear relationship between symptoms and TCM is of great significance for clinical practice and assisting physicians in diagnosis and treatment. Although there have been some studies on TCM prescription generation, these studies consider a single factor and directly model the symptom-prescription generation problem mainly based on symptom descriptions, lacking guidance from TCM knowledge. To this end, we propose a RoBERTa and Knowledge Enhancement model for Prescription Generation of Traditional Chinese Medicine (RoKEPG). RoKEPG is firstly pre-trained by our constructed TCM corpus, followed by fine-tuning the pre-trained model, and the model is guided to generate TCM prescriptions by introducing four classes of knowledge of TCM through the attention mask matrix. Experimental results on the publicly available TCM prescription dataset show that RoKEPG improves the $F_1$ metric by about 2\% over the baseline model with the best results.
\end{abstract}
\section{Introduction}
TCM is increasingly being used clinically worldwide, and its efficacy is recognized by more and more people and countries \cite{helfenstein2017analyzing}. TCM has accumulated a large body of literature and treatment records. A prescription consisting of TCM is the most critical form of TCM treatment, and it is widely used to treat various diseases. As of now, over 100,000 classical TCM prescriptions have been accumulated \cite{qiu2007traditional}, offering clinicians references and enabling computational models for computer-assisted therapy and automatic prescription generation through learning. In TCM prescription generation, herbs for treating symptoms are generated from symptom text. Recent deep learning breakthroughs enable mining complex TCM principles.

Although some studies on TCM prescription generation exist, such as PTM \cite{yao2018topic}, MC-eLDA \cite{zhang2019mc}, AttentiveHerb \cite{liu2019attentiveherb}, and TCM Translator \cite{wang2019tcm}, most of these studies only consider symptom factors to directly model the symptom-prescription generation problem, resulting in a lack of guidance of TCM knowledge for model generation. Moreover, there is a complex nonlinear relationship between the composition of TCM prescriptions and patient symptoms, and one herb may treat multiple symptoms, while one symptom may correspond to multiple herbs. Hence, it is challenging for traditional machine learning methods to fit the complex relationship between herbs and symptoms adequately. At the same time, thousands of herbs are covered in TCM prescriptions, and the unbalanced distribution of TCM data also poses a challenge regarding the accuracy of TCM prescription generation. Therefore, we take advantage of RoBERTa \cite{liu2019roberta} and propose a Knowledge Enhancement model for Prescription Generation of Traditional Chinese Medicine (RoKEPG) to address the above issues.

Our main contributions are as follows:
\begin{itemize}
	\item We performed a secondary pre-training of the RoBERTa model based on a TCM domain corpus which we constructed with 13.5 million words.
	\item We conduct fine-tuning on the pre-trained RoBERTa model based on the attention mask matrix to introduce the knowledge of nature, taste, channel tropism and effect of herbs for guiding the model to generate TCM prescriptions.
	\item The RoKEPG model is tested on a publicly available TCM prescription dataset by a series of experiments. The experimental results verify the effectiveness of the RoKEPG model compared with the state-of-the-art related models.
\end{itemize}
\section{Related Work}
This section reviews the work related to prescription generation for TCM. Early studies mainly relied on topic models for TCM prescription generation \cite{yao2018topic,zhang2019mc}. However, there are over 1000 commonly used herbs, and some of the correlations between herbs and symptoms are highly complex and challenging to compute, making topic models insufficient for the TCM prescription generation task. Subsequently, some scholars modeled TCM prescription generation as a Multilabel Classification (MLC) problem, considering each herb as a category label and predicting probabilities \cite{boutell2004learning,zhang2007ml,elisseeff2001kernel,read2011classifier}, but these methods did not fully utilize the correlations between labels. Inspired by Neural Machine Translation (NMT) models, research emerged on using Sequence to Sequence (Seq2Seq) models for generating TCM prescriptions \cite{liu2019attentiveherb,wang2019tcm,li2018exploration, li2020herb}.

In recent years, pre-trained language models have rapidly developed, and research on TCM prescription generation using pre-trained language models has also emerged \cite{liu2022novel,wangRe}. By harnessing the powerful semantic capabilities obtained from pre-training on large-scale unsupervised corpora, models can achieve good quality patient prescriptions with fine-tuning on small sample data. Additionally, methods based on heterogeneous graph contrastive learning have been used to generate TCM prescriptions \cite{yin2022hgcl}.

The above studies on TCM prescription generation mainly focused on directly modeling the symptom-TCM prescription generation problem based on symptoms. While some studies introduced the effects of herbs to guide prescription generation, these factors were relatively single. This inspired the current study to consider more influencing factors in the TCM prescription generation task, such as the nature, taste, and channel tropism of herbs, aiming to make the model generation process more in line with the realistic scenario of prescription writing by physicians.
\section*{Methods}
\subsection{Task Definition}
We first define the TCM prescription generation task, which is modelled based on the symptom text-prescription composition data pair, and finally achieve the generation of a set of herbs for treating symptom text. Specifically, TCM prescription generation refers to generating a set of herbs corresponding to the treatment symptoms $Y=\left[y_1,y_2\ldots,y_m\right]$ based on the symptoms $X=\left[x_1,x_2\ldots,x_n\right]$ of a TCM prescription, where $n$ is the length of the symptom text contained in the prescription and $m$ is the number of herbs contained in the prescription.
\subsection{Overview}
\begin{figure*}[t]
	\includegraphics[width=\textwidth]{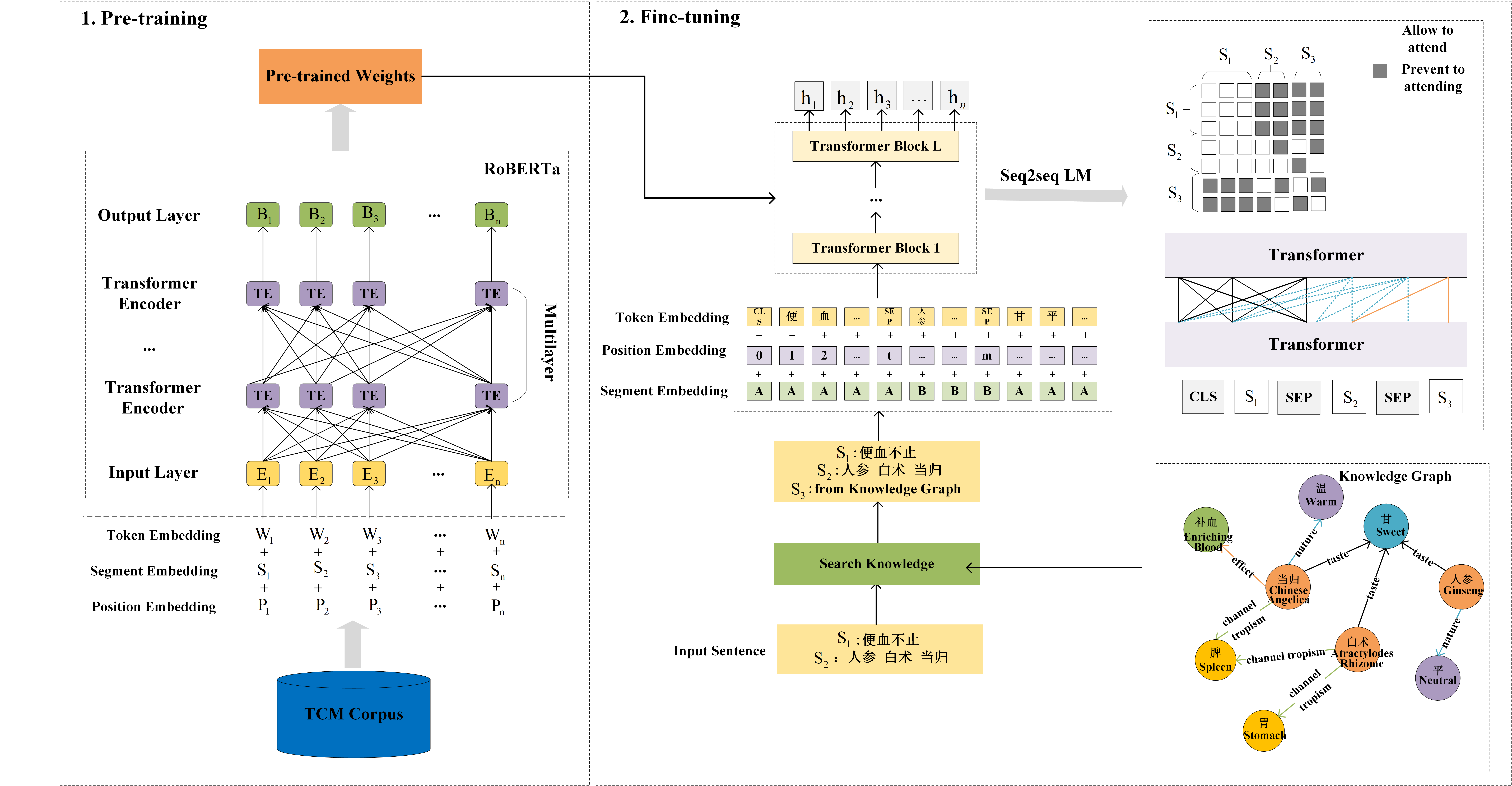}
	\caption{Framework diagram of RoKEPG model} 
	\label{frame}
\end{figure*}

The RoKEPG model consists of two main parts: pre-training and fine-tuning. In the pre-training stage, RoBERTa \cite{liu2019roberta}, pre-trained language model, is utilized and trained on the TCM corpus that we constructed. Subsequently, in the fine-tuning stage, the model generated during pre-training is used to generate TCM prescriptions. During fine-tuning, TCM knowledge is incorporated through the attention mask matrix to guide the model's generation process. The overall framework of the RoKEPG model is shown in Figure \ref{frame}, where the white squares in the fine-tuning phase indicate contextual information available when performing sequence-to-sequence learning, such as for setting symptom texts to do bidirectional attention when performing training. In contrast, grey squares indicate information that is not visible through the mask, such as for setting up the prediction of herbs to do unidirectional attention and for setting up knowledge that is not a counterpart of that herb to be invisible.
\subsection{Pre-training}

To improve the performance on the TCM prescription generation task, we pre-train the RoBERTa \cite{liu2019roberta} language model on a domain-specific TCM corpus. As publicly available pre-trained language models are generally based on general domain corpora, using them directly for domain-specific tasks may not yield satisfactory results. Therefore, we constructed a TCM corpus containing 13.5 million words by processing relevant TCM books like Basic Theory of Traditional Chinese Medicine \cite{sunBasic} and Pharmacology of Traditional Chinese Medical Formulae \cite{dengFang}. After constructing the TCM corpus, we proceed to pre-train the model using this corpus, as shown in Figure \ref{frame}.

The input of the RoBERTa model contains three parts: Token Embedding, Segment Embedding and Position Embedding. The three parts are summed and fed into the architecture containing the L-layer Transformer. For each layer $l$, the Self-Attention Head is used to aggregate the output from the previous layer and learn the contextual representation $H$. The formula for the Self-Attention Head $A_l$ is shown below:
\begin{equation}
	\begin{cases}Q=H^{l-1}W_l^Q\\K=H^{l-1}W_l^K\\V=H^{l-1}W_l^V&\end{cases}
\end{equation}
\begin{equation}
	A_l=softmax\left(\frac{QK^T}{\sqrt{d_k}}\right)V_l
\end{equation}
where $Q$ is the query matrix, $K$ is the key matrix, $V$ is the value matrix, $H^{l-1}$ is the output vector from layer $l-1$, and $W_l^Q$, $W_l^K$, and $W_l^V$ are the parameter matrices.
\subsection{Fine-tuning}
\subsubsection{Overall Description of the Fine-tuning Phase}
UniLM \cite{dong2019unified} is a unified pre-trained language model with natural language understanding (NLU) and natural language generation (NLG) capabilities, which enables the model to implement Seq2Seq through Attention Mask. Inspired by the UniLM model, we enable the RoBERTa model to implement Seq2Seq learning by changing the attention matrix masks. Specifically, based on the model weights obtained in the pre-training phase, mask fine-tuning on RoBERTa to generate TCM prescriptions. The objective function of the TCM prescription generation task:
\begin{equation}
	\arg\max_{y\in V}P(Y|X)
\end{equation}
\begin{equation}
	P(Y|X,\theta)=\prod_{t=1}^mp(y_t|y_{t-1},y_{t-2},\ldots,y_1;X;\theta)
\end{equation}
where $\theta$ is the parameter for model training, $V$ is the dictionary corresponding to the model, and the goal of model training is to maximize the joint probability of generating TCM prescriptions under the condition that the symptoms are $X$.

Since the original RoBERTa model predicts individual words by masking, this masking approach cannot capture the complete semantic relationships of TCMs, and at the same time, increases the difficulty of the model prediction. Therefore, we choose to perform the mask prediction for the full name of the Chinese medicine in the composition of the TCM prescription.
\subsubsection{Input Layer}
Firstly, the symptom and prescription composition corresponding to each TCM prescription in the dataset is used as a sequence to the source sequence $X$ and the target sequence $Y$ in the sequence task, respectively. The symptom and prescription composition sequences $<X,Y>$ are concatenated with the special characters "[CLS]" and "[SEP]" to perform splicing:
\begin{equation}
	S=\left(\left[CLS\right],X,\left[SEP\right],Y,\left[SEP\right]\right)
	\label{eq}
\end{equation}
where "[CLS]" indicates the beginning of the sequence and "[SEP]" is used to separate the source sequence from the target sequence as well as to indicate the end of the target sequence. To distinguish the source and target sequences in the input text, we set the segment embedding of each character in the symptom sequence $X$ to $A$, and the segment embedding of each character in the prescribed constituent sequence $Y$ to $B$.

\begin{figure}
	\centering
	\includegraphics[width=0.5\linewidth]{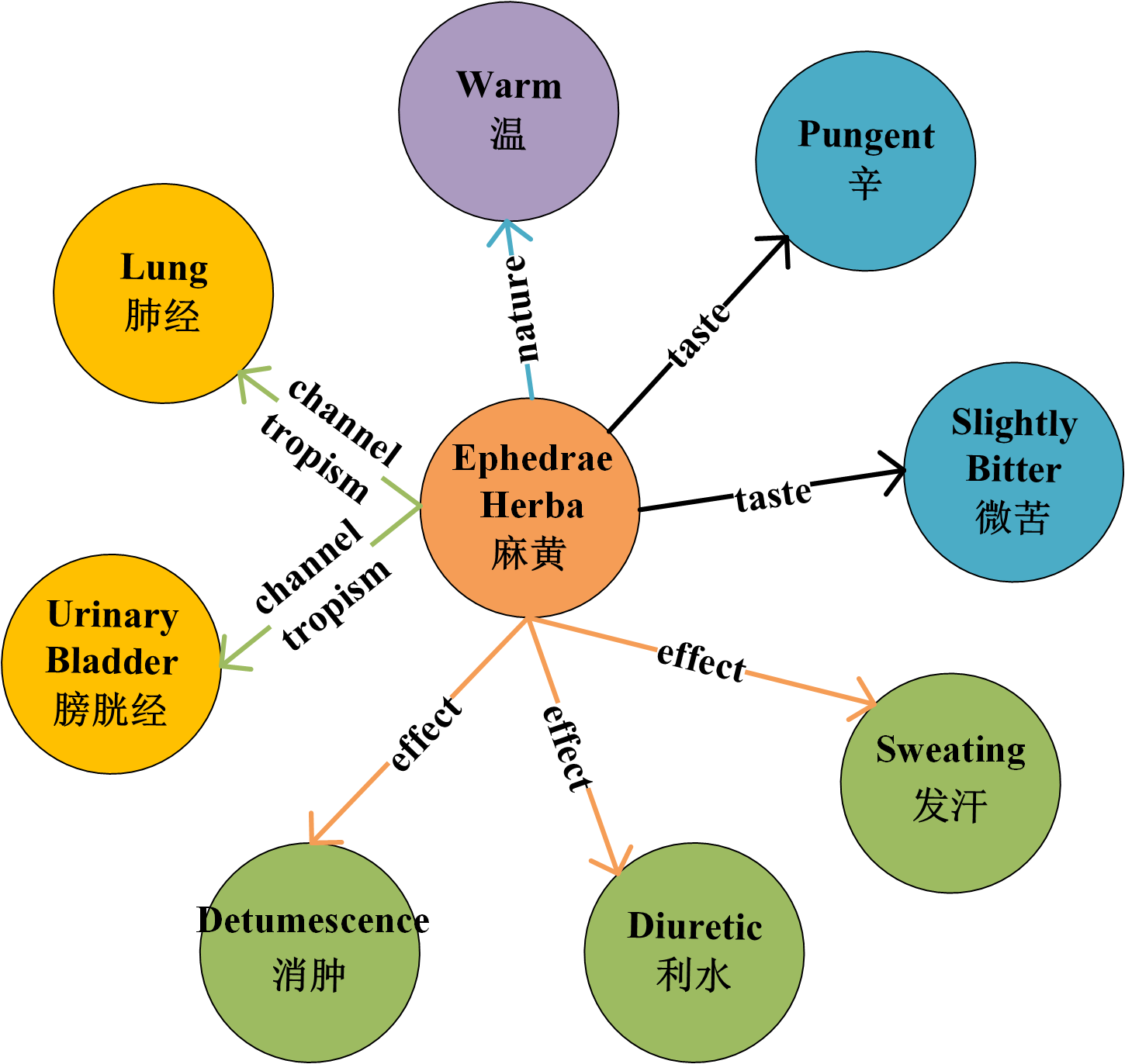}
	\caption{Diagram of TCM Knowledge Graph} 
	\label{diag}
\end{figure}

TCM prescription generation is not only related to symptoms but also closely related to TCM knowledge. Therefore, we introduce some additional TCM knowledge to guide the model generation. Firstly, a TCM-centric knowledge graph is constructed, containing five categories of nodes: herbs, nature, taste, channel tropism and effects, taking Ephedrae Herba as an example, as shown in Figure \ref{diag}.

\begin{figure}[]
	\includegraphics[width=\columnwidth]{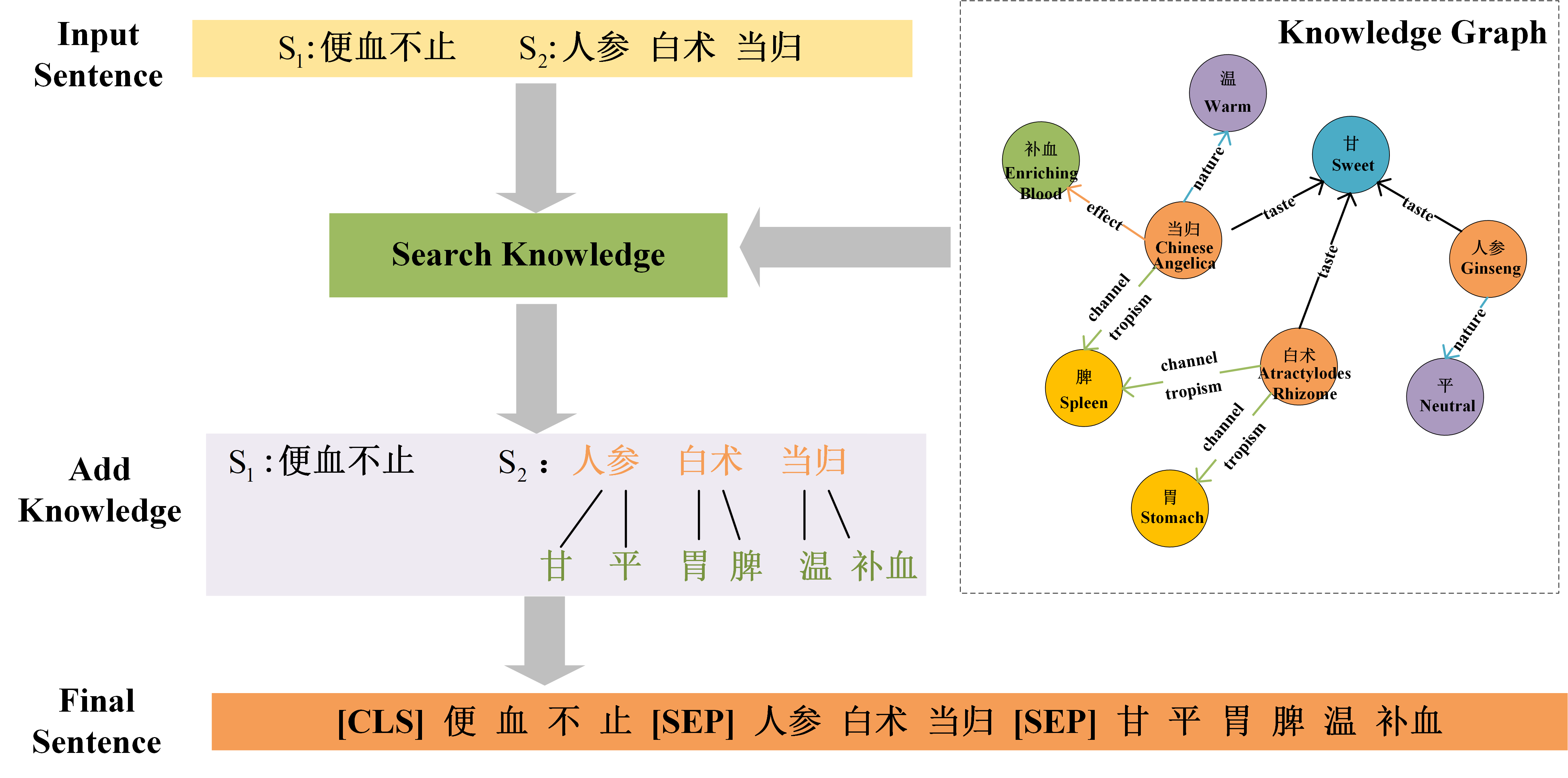}
	\caption{Prescription composition adding knowledge example diagram} 
	\label{Prescription}
\end{figure}

During the model training, the input sequence of the model needs to add knowledge of TCM. Take the TCM prescription with the symptom of "便血不止(bleeding in the stool)" and the component of the prescription "人参、白术、当归(Ginseng, Largehead Atractylodes Rhizome and Chinese Angelica)" as an example, extract the triples of each herb in the prescription component "人参、白术、当归(Ginseng, Largehead Atractylodes Rhizome and Chinese Angelica)". The ternary groups of the corresponding entities in the knowledge graph are added to the input sequence, as shown in Figure \ref{Prescription}.

Equation \ref{eq} is correspondingly modified to:
\begin{equation}
	S=\left(\left[CLS\right],X,\left[SEP\right],Y,\left[SEP\right],\ Z\right)
\end{equation}
where $Z$ denotes the sequence consisting of the relevant knowledge of each herb in $Y$.

\begin{figure}[]
	\includegraphics[width=\columnwidth]{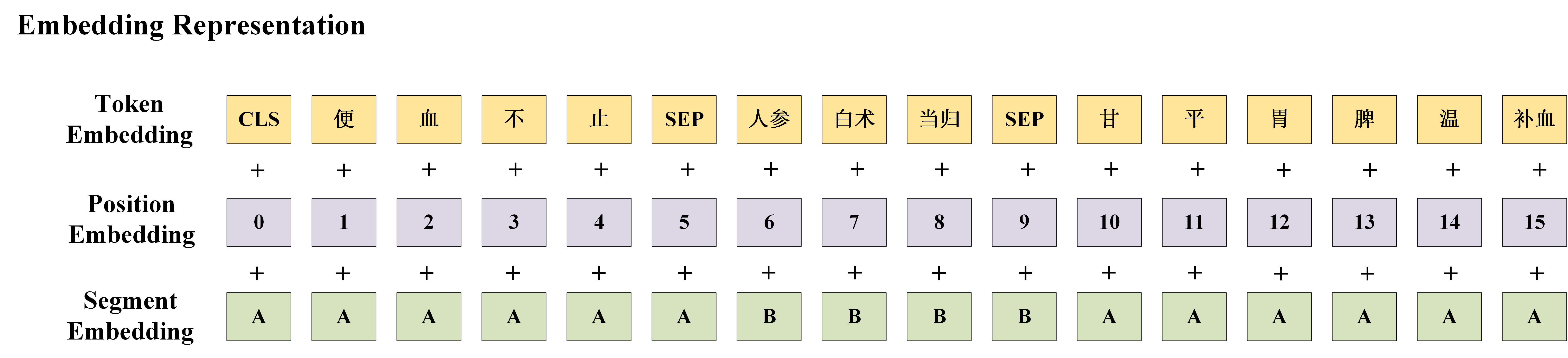}
	\caption{Embedding calculation diagram} 
	\label{embedding}
\end{figure}

The segment embedding corresponding to the joined TCM knowledge is set to $A$ to distinguish it from the prescription component. The token embedding, position embedding and segment embedding of the spliced sequence $S$ are summed and input to the RoBERTa model as the final input vector, and the embedding of each character is calculated as shown in Figure \ref{embedding}.

Map the input text sequence $S$ to the vector sequence $H^0$ correspondingly:
\begin{equation}
	h_n^0=E^{emb}\left(s_n\right)+E^{pos}\left(s_n\right)+E^{seg}\left(s_n\right),n\in\left[1,N\right]
\end{equation}
\begin{equation}
	H^0=\left[h_1^0,h_2^0,\ldots,h_N^0\right]
\end{equation}
\subsubsection{Mask Prediction Layer}
The $H^0$ in the input layer is fed into the network consisting of the L-layer Transformer to obtain the contextual representation of the input text at the lth layer:
\begin{equation}
	H^l\ =\ {\rm Transformer}_l\left(H^{l-1}\right),l\in\left[1,L\right]
\end{equation}

The original RoBERTa model utilizes contextual information from both the left and right sides of a character when making predictions, which is not directly applicable to sequence-to-sequence tasks. To enable the RoBERTa model for sequence-to-sequence learning, we regulate the contextual information that characters in a sequence can access through attention masks. Specifically, for the TCM prescription generation task, characters in the symptom text component need to have visibility to information from both the left and right sides. On the other hand, for characters in the prescription component, only the information of the first $j-1$ characters is accessible when predicting the $j^{th}$ character.

\begin{figure}[]
	\centering
	\includegraphics[width=0.9\linewidth]{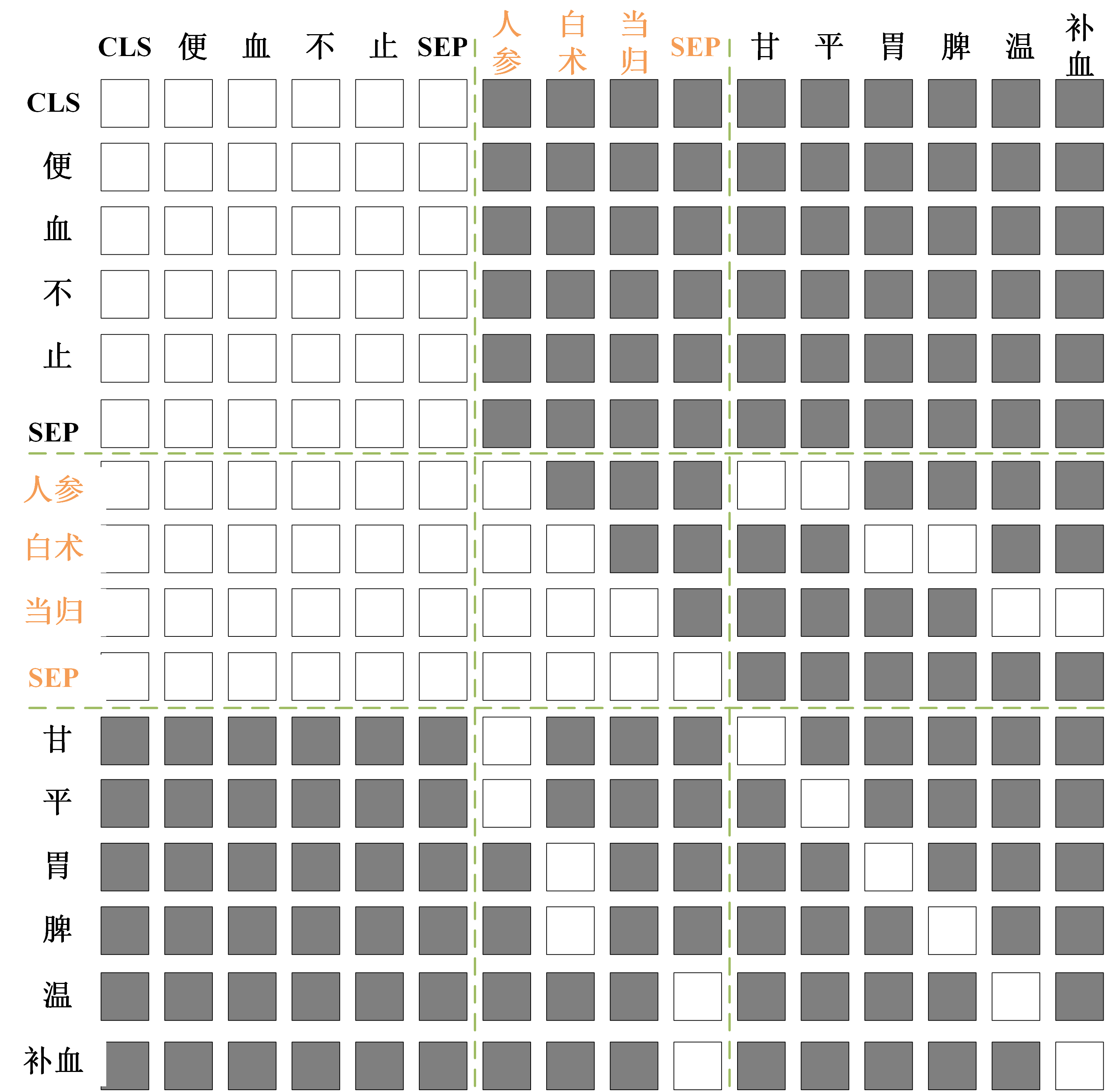}
	\caption{Mask matrix diagram} 
	\label{Mask}
\end{figure}

During model training, with the introduction of additional TCM knowledge in the input layer, the prescription component needs to access the knowledge of a specific herb when making predictions, while excluding knowledge of other herbs. This necessitates refining the attention mask matrix to precisely control the accessible information.Taking the Chinese prescription with the symptom component ”便血不止(blood in stool)” and the target prescription component ”人参、白术、当归(Ginseng, Largehead Atractylodes Rhizome and Chinese Angelica)” as an example, the mask matrix for introducing the knowledge of Chinese medicine is shown in Figure \ref{Mask}.

Specifically, the symptom part of the attention mask matrix is set to 0, meaning that all symptom sequence characters are involved in the self-attention calculation. The right part of the characters of the TCM prescription component that needs to be predicted is set to $-\infty$, while the mask of the Chinese medicine corresponding to the Chinese medicine knowledge that needs to be predicted is set to 0, and the mask of the knowledge corresponding to other Chinese medicines is set to $-\infty$, indicating that the Chinese medicine can see its own related knowledge but not other irrelevant knowledge. With such a setting, all the characters of the symptom sequence can be made to perform self-attentive computation, and the left part of the predicted characters of the prescription component and the corresponding Chinese medicine knowledge part participate in the attention computation, while its right part does not participate in the computation. The attention head $A_l$ is calculated by the formula:
\begin{equation}
	\begin{cases}Q=H^{l-1}W_l^Q\\K=H^{l-1}W_l^K\\V=H^{l-1}W_l^V\end{cases}
\end{equation}
\begin{equation}
	M_{ij}=\left\{\begin{matrix}0,&\text{allow to attend}\\-\infty,&\text{prevent from attending}\end{matrix}\right.
\end{equation}
\begin{equation}
	A_l=\text{softmax}\left(\frac{QK^T}{\sqrt{d_k}}+M\right)V_l
\end{equation}
\subsubsection{Output Layer}
To output a set of herbs, the end of a multilayer Transformer is connected to a linear classifier (such as a softmax classifier) to generate a probability distribution over the word list to generate a prescription for the herbs:
\begin{equation}
	P\left(y\right)=softmax\left(h_{\left[Mask\right]}^LW^T\right)
\end{equation}

The predicted herb is appended to the input sequence to replace the $\left[MASK\right]$ tokens to continue decoding until $\left[SEP\right]$ appears. The cross-entropy loss $L$ of the whole target sequence $Y$ is defined as:

\begin{equation}
	L=-\frac1{|Y|}\sum_{i=1}^{|Y|}\log p\left(y_i|y_{<i},X\right)
\end{equation}
where $Y$ is the target sequence of the TCM prescription, $y_i$ is the ith herb generated.

We view TCM prescription generation as a text generation problem in which the choice of decoding method significantly affects the quality of the generated text. During the model's test inference phase, we refrain from using the common Beam Search algorithm due to its drawbacks in causing duplication and generating shorter texts, which are not ideal for TCM prescriptions. Instead, we employ a combined approach using Temperature-controlled Stochastic Sampling and Top-p Sampling as our decoding methods. This combination enhances the RoKEPG model's prescription generation by considering the diverse nature of TCM prescriptions. As TCM prescriptions are composed of a varied set of herbs without repetition and vary in length, our decoding strategy aims to improve both the diversity and accuracy of the generated prescriptions, ensuring better results for practical applications.
\section{Experiment}
In this section, the RoKEPG model is systematically evaluated on a publicly available TCM prescription dataset.
\subsection{Dataset}
\textbf{TCM Corpus}：We collected books from more than twenty books related to the field of TCM, such as Basic Theory of Traditional Chinese Medicine \cite{sunBasic} and Pharmacology of Traditional Chinese Medical Formulae \cite{dengFang}. Finally, after cleaning and organising the data, we constructed a TCM corpus containing 340,000 lines and 13.5 million words.

\textbf{TCM Knowledge Graph}：In TCM theory, herbs have properties such as nature, taste, channel tropism and effect. We construct a knowledge graph centred on TCM, which contains five types of entities (TCM, nature, taste, channel tropism, effect) with a total number of 1430 entities and four types of relationships (TCM-has\_nature$\rightarrow$nature, TCM-has\_taste$\rightarrow$taste, TCM-has\_channel$\rightarrow$channel tropism, TCM-has\_effect$\rightarrow$ effect) with a total number of 9853 relationships.

\textbf{TCM Dataset}：The publicly available dataset contains 82,044 Traditional Chinese Medicine (TCM) prescriptions, obtained from traditional prescription records in TCM books, and underwent data cleaning and collation. We randomly divide the whole TCM dataset into three parts: train set (73,839 entries, about 90\%), valid set (4,102 entries, about 5\%) and test set (4,103 entries, about 5\%).

\subsection{Evaluation Metrics and Experimental Setup}
\textbf{Evaluation Metrics}：Adopt the practice with other papers of TCM prescription generation by Seq2Seq \cite{li2018exploration}, we used the same Precision, Recall and $F_1$ scores as metrics to evaluate the results of TCM prescription generation, with Precision and Recall calculated as follows:
\begin{equation}
	Precision=\frac{G\left(X\right)\cap R\left(X\right)\mathrm{\ } }{\mathrm{\ G(X)\ }}
\end{equation}
\begin{equation}
	Recall=\frac{G\left(X\right)\cap R\left(X\right)\mathrm{\ } }{\mathrm{\ R(X)\ }}
\end{equation}
Where $G(X)$ is a set of herbs generated by the model and $R(X)$ is the prescription component in the reference prescription. In the TCM prescription generation task, Recall is more critical than Precision \cite{li2020herb}. The higher the Recall, the more the generated set of herbs also belong to the real prescription composition. However, in order to avoid generating too long prescriptions, it is also necessary to consider Precision, so the prescription generation results are evaluated comprehensively using the $F_1$ metric because $F_1$ takes into account the Precision and Recall are considered together.
\begin{equation}
	F_1=\frac{2\cdot\mathrm{\ Precision\ } \cdot\mathrm{\ Recall\ } }{\mathrm{\ Precision\ }+\mathrm{\ Recall\ }}
\end{equation}

\textbf{Experimental Setup}: Since almost all TCM prescription compositions in the TCM dataset contain 20 or fewer herbs, we set the maximum number of generated prescriptions containing herbs to 20. All experiments were performed in PyTorch, and we set the maximum length of the input sequence to 256 and selected the best experimental results based on the valid set.

In the pre-training phase of RoKEPG, the number of hidden layer neurons is set to 768, the number of hidden layers is set to 12, the head of multi-head attention is set to 12, the batch size is 32, the number of epochs is 15, the maximum sequence length is set to 512, and the secondary pre-training is based on the RoBERTa-WWM-Chinese model.

In the fine-tuning phase, the number of hidden layer neurons is set to 768, the number of hidden layers is set to 6, the head of multi-head attention is set to 12, the batch size is 16, the maximum sequence length is set to 256, the learning rate is 1e-5, the number of epochs is 40, and the model is fine-tuned based on the model obtained in the pre-training phase.

In the test decoding stage, the p-value of Top-p Sampling is set to 0.8, while the temperature is set to 0.2.
\subsection{Comparative Experiments}
To validate the RoKEPG model, we compared it with the following baseline models on the same dataset:
\begin{itemize}
	\item	Multi-Label\cite{li2018exploration}: The encoder part uses BiGRNN, and the generation part uses a multi-label classification method to predict the model of Chinese medicine.
	\item	Basic Seq2Seq\cite{li2018exploration}: The encoder is a bidirectional Gated Recurrent Unit (GRU), and the decoder is a unidirectional GRU, forming a sequence-to-sequence model.
	\item	S2SA-CS\cite{li2018exploration}: A Seq2Seq model with attention mechanism, coverage mechanism and soft cross-entropy.
	\item	TCM-Translator\cite{wang2019tcm}: The encoder is a Transformer without positional embedding, and the decoder is an LSTM, constituting the Seq2Seq model.
	\item	Herb-Know\cite{li2020herb}: A sequence-to-sequence model based on pointer networks and improved model generation by introducing knowledge of the effects of herbs.
	\item	BERT-UniLM\cite{wangRe}: A single BERT generates TCM prescriptions by attention mask matrix.
	\item	RoBERTa-UniLM\cite{liu2019roberta}: A single RoBERTa is implemented to generate TCM prescriptions through an attention mask matrix similar to BERT-UniLM.
	\item	GPT-2\cite{radford2019rewon}: An autoregressive model consisting of the decoder part of a multilayer one-way Transformer. 

	\item	BART\cite{lewis2019bart}: A pre-trained language model, built according to Transformer's general framework, implements sequence-to-sequence with a bidirectional encoder and a left-to-right autoregressive decoder.
	\item	T5\cite{raffel2020exploring}: A pre-trained language model that retains most of the original Transformer framework transforms all tasks in natural language processing into text-to-text tasks.
\end{itemize}

Through several experiments, the learning rates of GPT-2, BART and T5 are all set to 1e-5, the batch sizes are all set to 8, and the number of epochs are all set to 30. the main parameters of BERT-UniLM and RoBERTa-UniLM are the same as those of RoKEPG, while Multi-Label, Basic Seq2Seq and S2SA-CS keep the same parameter settings as paper \cite{li2018exploration}, TCM-Translator keeps the same parameter settings as the original paper \cite{wang2019tcm}, and Herb-Know likewise keeps the same parameter settings as the original paper \cite{li2020herb}.
\subsubsection{TCM Prescription Generation Comparison Experiment}
The experimental results of TCM prescription generation based on the publicly available TCM dataset are shown in Table \ref{table:comparison}.

\begin{table}[]
	\renewcommand{\arraystretch}{1.2}  
	\begin{center}
		\caption{TCM prescription generation comparison experiment}
		\label{table:comparison}
		\begin{tabular}{cccc}
			\hline
			Model          & Precision & Recall & $F_1$           \\ \hline
			Multi-Label    & 10.83     & 29.72  & 15.87          \\
			Basic Seq2Seq  & 26.03     & 13.52  & 17.80          \\
			S2SA-CS        & 29.57     & 17.30  & 21.83          \\
			TCM-Translator & 25.72     & 12.13  & 16.48          \\
			Herb-Know      & 28.35     & 20.56  & 23.83          \\
			GPT-2          & 15.47     & 13.24  & 14.27          \\
			BART           & 20.56     & 14.84  & 17.24          \\
			T5             & 19.34     & 16.98  & 18.08          \\
			BERT-UniLM     & 28.82     & 18.85  & 22.79          \\
			RoBERTa-UniLM  & 27.76     & 19.46  & 22.88          \\
			RoKEPG         & 23.43     & 29.01  & \textbf{25.92} \\ \hline
		\end{tabular}
	\end{center}
\end{table}
As can be seen from the experimental results in Table \ref{table:comparison}, from the overall experimental results, the RoKEPG model significantly outperforms the other baseline models in terms of recall and $F_1$ scores, and the RoKEPG model outperforms the most effective baseline Herb-Know model by about 2\% in the $F_1$ metric.

The Multi-Label model has a high Recall but low Precision and $F_1$ values due to the long prescription length generated by the Multi-Label model. The results generated based on BERT are slightly lower than those generated based on RoBERTa, because the RoBERTa model is a tuned and enhanced version of BERT and does achieve better results than BERT in several NLP tasks. Although GPT-2, BART and T5 models are more widely used in text generation tasks, they do not work well in TCM prescription generation tasks.

\subsubsection{TCM Knowledge Introduction Experiment}
For the TCM knowledge introduction module, we explored two different introduction methods: random introduction and all introduction. The random introduction refers to finding all the triads associated with a certain TCM from the TCM knowledge graph and randomly selecting one of the triads as the TCM knowledge to be added to the input sequence. All introduction refers to adding all the triads associated with herb into the input sequence. We validate the effect of these two different knowledge introduction methods based on three models, BERT-UniLM, RoBERTa-UniLM, and RoKEPG together, and the experimental results are shown in Table \ref{table:ways}.
\begin{table}[]
	\renewcommand{\arraystretch}{1.2}  
	\begin{center}
		\caption{Different ways of introducing knowledge}
		\label{table:ways}
		\begin{tabular}{ccccc}
			\hline
			Knowledge Methods       & Model         & Precision & Recall & $F_1$           \\ \hline
			\multirow{3}{*}{Random Introduction} & BERT-UniLM    & 24.01     & 22.82  & 23.40          \\
			& RoBERTa-UniLM & 20.35     & 28.01  & 23.57          \\
			& RoKEPG        & 21.35     & 32.25  & 25.69          \\ \hline
			\multirow{3}{*}{All Introduction}    & BERT-UniLM    & 22.72     & 24.13  & 23.40          \\
			& RoBERTa-UniLM & 20.75     & 27.32  & 23.58          \\
			& RoKEPG        & 23.43     & 29.01  & \textbf{25.92} \\ \hline
		\end{tabular}
	\end{center}
\end{table}

The experimental results in Table \ref{table:ways} show that all three models are better with the all introduction approach than the random introduction approach, probably because there is a repeated introduction of the same relevant knowledge of Chinese medicine if the random introduction is performed in the training phase. Moreover, the random introduction only introduces one knowledge at a time, and the model receives less knowledge than the all introduction approach, leading to decreased model performance.
\subsubsection{Different Decoding Strategies}
In the decoding generation stage of the model test, different decoding strategies will affect the results differently. We combine a mixture of two decoding strategies, Temperature-controlled Stochastic Sampling and Top-p Sampling, for decoding. The results of comparing different decoding methods on the performance of RoKEPG model are shown in Table \ref{table:decoding}, and the contents in parentheses in the table are the parameter settings when the results of corresponding decoding methods are optimal.
\begin{table}[]
	\renewcommand{\arraystretch}{1.2}  
	\begin{center}
		\caption{Different decoding methods}
		\label{table:decoding}
		\begin{tabular}{cccc}
			\hline
			Decoding Methods         & Precision & Recall & $F_1$           \\ \hline
			Beam Search(beam-size=5) & 23.09     & 27.97  & 25.29          \\
			Top-K (k=3)              & 21.26     & 29.50  & 24.71          \\
			Top-P (p=0.9)            & 23.64     & 28.65  & 25.90			 \\ 
			Temperature(temperature=0.2) & 23.00 & 28.91 & 25.62            \\
			RoKEPG (mixed decoding)      & 23.43 & 29.01 & \textbf{25.92}		\\ \hline
		\end{tabular}
	\end{center}
\end{table}

The experimental results in Table \ref{table:decoding} show that the mixture decoding method of RoKEPG model outperforms the other decoding methods, and the effect of Temperature-controlled Stochastic Sampling and Top-P Sampling alone is slightly inferior to that of the mixture decoding of RoKEPG model. The common Beam Search decoding method is biased to generate shorter prescriptions, and there are many duplicates in the results compared with other decoding methods.

\subsubsection{Ablation Experiment}
In order to explore the effects of the pre-training phase and the introduction of TCM knowledge in the RoKEPG model, we designed two sets of ablation experiments.
We first validate the effectiveness of the pre-training phase based on the RoKEPG model and the RoBERTa-UNILM model together, and the experimental results are shown in Table \ref{table:ablation}. w/o denotes Without, and the pre-training module is denoted by pre-trained.
\begin{table}[]
	\renewcommand{\arraystretch}{1.2}  
	\begin{center}
		\caption{Ablation experiments with pre-training module}
		\label{table:ablation}
		\begin{tabular}{cccc}
			\hline
			Model                         & Precision & Recall & $F_1$    \\ \hline
			RoBERTa-UNILM                 & 24.18     & 25.75  & \textbf{24.94} \\
			RoBERTa-UNILM-w/o pre-trained & 27.76     & 19.46  & 22.88 \\
			RoKEPG                        & 23.43     & 29.01  & \textbf{25.92} \\
			RoKEPG-w/o pre-trained        & 20.75     & 27.32  & 23.58 \\ \hline
		\end{tabular}
	\end{center}
\end{table}

As shown from the experimental results in Table \ref{table:ablation}, the pre-trained model significantly improves effectiveness, with more than 2\% improvement in the $F_1$ metric over the non-pre-trained model.

The results of the ablation experiment for the effectiveness of introducing TCM knowledge into the RoKEPG model are shown in Table \ref{table:ablation2}, and the introduction of TCM knowledge is indicated by herb-know.
\begin{table}[]
	\renewcommand{\arraystretch}{1.2}  
	\begin{center}
		\caption{Ablation experiments for knowledge introduction}
		\label{table:ablation2}
		\begin{tabular}{cccc}
			\hline
			Model                & Precision & Recall & $F_1$    \\ \hline
			RoKEPG               & 23.43     & 29.01  & \textbf{25.92} \\
			RoKEPG-w/o herb-know & 24.18     & 25.75  & 24.94  \\ \hline
		\end{tabular}
	\end{center}
\end{table}

From the experimental results in Table \ref{table:ablation2}, it can be seen that the introduction of TCM knowledge brings a 0.98\% improvement in the $F_1$ index compared with the non-introduction of TCM knowledge, indicating that the introduction of TCM knowledge does help in TCM prescription generation, and the model enhanced by TCM knowledge can generate TCM prescriptions better.
\subsubsection{Case Study}
Further, we randomly selected two TCM prescriptions from the test set for case study, as shown in Table \ref{table:case}. The RoKEPG model generates herbs mostly present in the reference prescription, including some not originally listed.

\begin{table*}[t]
	\caption{Case study}
	\label{table:case}
	\renewcommand{\arraystretch}{1.25}  
	\begin{tabularx}{\textwidth}{p{0.08\textwidth}p{0.45\textwidth}p{0.4\textwidth}}
		\hline
		Cases                  & Case 1 & Case 2 \\
		\hline
		Symptom                & 小儿乳癖、积聚。按之苦痛，肌肤渐瘦，面色青黄；小儿阴阳气不顺，虚痞胀满，呕逆腹痛，成癥瘕痞结。 \newline (Lactation and accumulation in children. Bitter and painful when pressed, skin gradually thinned, face green and yellow; Yin and Yang Qi disorder in children, deficiency and fullness, vomiting and rebellious abdominal pain, forming abdominal masses and abnormalities.) & 产后赤白痢，脐下咻痛。\newline (Postpartum red and white dysentery with painful swoosh under the umbilicus.) \\
		\hline
		Reference prescription & 麦芽，灵脂，香附，陈皮，神曲，青皮，莱菔，厚朴，半夏，槟榔，枳实，砂仁，三棱。 \newline (Hordei Fructus Germinatus, Trogopterus Dung, Nutgrass Galingale Rhizome, Citrus reticulata Blanco, Medicated Leaven, Citri Reticulatae Pericarpium Viride, Radish root, Magnoliae Officmalis Cortex, Pinellia ternate, Betelnutpalm Seed, Immature Bitter Orange, Villous Amomum Fruit, Sparganii Rhizoma.) & 人参，白术，麦冬，当归，熟地，天麻，防风，荆芥，陈皮，甘草，生姜。\newline (Ginseng, Largehead Atractylodes Rhizome, Radix Ophiopogonis, Chinese Angelica, Rehmanniae Radix Praeparata, Tall Gastrodia Tuber, Divaricate Saposhniovia Root, Schizonepetae Herba, Citrus reticulata Blanco, Licorice, Fresh Ginger.) \\
		\hline
		RoKEPG                 & 枳实，陈皮，厚朴，香附，山楂，麦芽，神曲，砂仁，木香，槟榔， 青皮，甘草。 \newline (Immature Bitter Orange, Citrus reticulata Blanco, Magnoliae Officmalis Cortex, Nutgrass Galingale Rhizome, Crataegi Fructus, Hordei Fructus Germinatus, Medicated Leaven, Villous Amomum Fruit, Common Vladimiria Root, Betelnutpalm Seed, Citri Reticulatae Pericarpium Viride, Licorice.) & 人参，白术，当归，川芎，白芍，熟地，麦冬，陈皮，甘草，生地黄。 \newline (Ginseng, Largehead Atractylodes Rhizome, Chinese Angelica, Sichuan lovase rhizome, White Peony Root, Rehmanniae Radix Praeparata, Radix Ophiopogonis, Citrus reticulata Blanco, Licorice, Rehmanniae Radix.) \\
		\hline
		Result                 & Precision=66.67 Recall=61.54 $F_1$=64.0 & Precision=70.0 Recall=63.64 $F_1$=66.67 \\
		\hline
	\end{tabularx}
\end{table*}
In the first case, the RoKEPG model generated several herbs that were not in the reference prescription according to the symptoms, including Crataegi Fructus for treating gastric and epigastric fullness, diarrhea and abdominal pain \cite{dongShan}; the generated Common Vladimiria Root is mainly used for symptoms such as epigastric distension and inappetence \cite{zhengmu}; and the generated Citri Reticulatae Pericarpium Viride could be used for treating lactation, canker sores, and epigastric distension and pain \cite{xuZhong}. The symptoms these herbs can treat are similar to those of the first case, indicating that the RoKEPG model is not randomly generated herbs, as all the herbs generated by the model are highly correlated with the prescribed symptoms.

In the second case, the RoKEPG model generated Sichuan lovase rhizome and White Peony Root, which is one of the well-known pairs of medicines that can invigorate blood and nourish it and soften the liver to relieve pain \cite{wuJi}. Red and white dysentery (赤白痢) refers to the presence of pus and blood in the stool often accompanied by abdominal pain. The medicine pair composed of Sichuan lovase rhizome and White Peony Root can relieve the pain. The model generated Rehmanniae Radix, which also treats blood in the stool and stops bleeding \cite{yeSheng}. Based on the above analysis, the RoKEPG model can learn the combination and pairs of Chinese medicines.
\section{Conclusion}
TCM is an essential part of traditional medicine in China, and it is crucial to study the prescription generation of TCM and to uncover the complex relationship between herbs and symptoms. For the TCM prescription generation task, we propose a RoBERTa and Knowledge Enhancement model for Prescription Generation of Traditional Chinese Medicine model (RoKEPG). RoKEPG performes a secondary pre-training of the RoBERTa model based on a TCM domain corpus which we constructed, and introduces four types of knowledge of TCM through the attention mask matrix to help the model to generate prescriptions better. The experimental results based on the publicly available TCM dataset show that RoKEPG improves the $F_1$ index by about 2\% over the best baseline model, which can effectively assist TCM practitioners in prescribing. Since we model the TCM prescription generation as a text generation problem, there is the problem of exposure bias in the text generation task, and we will also consider adding contrastive learning to alleviate this problem in the follow-up to improve the quality of the generated TCM prescriptions.
\section*{Acknowledgement}
This work was supported by National Key R\&D Program of China (2019YFC1711000) and Collaborative Innovation Center of Novel Software Technology and Industrialization.

\bibliographystyle{IEEEtran}
\bibliography{refs.bib}
\end{CJK}
\end{document}